\documentclass{article}

\usepackage{arxiv}

\usepackage[utf8]{inputenc} % allow utf-8 input
\usepackage[T1]{fontenc}    % use 8-bit T1 fonts
\usepackage{hyperref}       % hyperlinks
\usepackage{url}            % simple URL typesetting
\usepackage{booktabs}       % professional-quality tables
\usepackage{amsfonts}       % blackboard math symbols
\usepackage{nicefrac}       % compact symbols for 1/2, etc.
\usepackage{microtype}      % microtypography
\usepackage{lipsum}		% Can be removed after putting your text content
\usepackage{graphicx}
\usepackage{doi}

\usepackage{graphicx}
\usepackage{mathptmx}      
\usepackage{amsfonts}
\usepackage{todonotes}      
\usepackage{subfig}
\usepackage{hyperref}
\usepackage{multicol}
\usepackage{adjustbox}

\title{Multi-Branch Deep Radial Basis Function Networks for Facial Emotion Recognition\thanks{This work was supported by project grant CONACYT CB-S-26314. The first author is supported by scholarship No. CONACYT  731653. This preprint has not undergone peer review or any post-submission improvements or corrections. The Version of Record of this article is published in Neural Computing and Applications, and is available online at https://doi.org/10.1007/s00521-021-06420-w}}

\date{} 					% Or removing it

\author{Fernanda Hern\'andez-Luquin\\ 
        \texttt{marialuquin@inaoep.mx}
	\And
	 Hugo Jair Escalante\\
    \texttt{hugojair@inaoep.mx}
	}

\renewcommand{\shorttitle}{\textit{arXiv} Template}

\begin{document}
\maketitle

\begin{abstract}
Emotion recognition (ER) from facial images is one of the landmark tasks in affective  computing with major developments in the last decade. Initial efforts on ER relied on handcrafted features that were used to characterize facial images and then feed to standard predictive models. Recent methodologies comprise end-to-end trainable deep learning methods that simultaneously learn both, features and predictive model.  Perhaps the most successful models are based on  convolutional neural networks (CNNs). While these models have excelled at this task, they still fail at capturing local patterns that could emerge in the learning process. We hypothesize these patterns could be captured by variants based on locally weighted learning.  
Specifically, in this paper we propose a CNN based architecture enhanced with multiple branches formed by radial basis function (RBF) units that aims at exploiting local information at the final stage of the learning process. Intuitively, these RBF units  capture local patterns shared by similar instances using an intermediate representation, then the outputs of the RBFs are feed to a softmax layer that exploits this information to improve the predictive performance of the model. This feature could be particularly advantageous in ER as cultural / ethnicity differences may be identified by the local units.   We evaluate the proposed method in several ER datasets and show  the proposed methodology achieves state-of-the-art in some of them, even when we adopt a pre-trained VGG-Face model as backbone. We show it is the incorporation of local information what makes the proposed model competitive. 


\end{abstract}

\section{Introduction}
\label{sec:intro}
Automated emotion Recognition (ER) is the ability to identify human emotional states by analyzing speech, facial expressions, and body gestures~\cite{liang2018multimodal}. ER has been proved to be very useful in areas as: affective computing, human-computer interaction, and as support tool for psychology, psychiatry, neurology, and related applications and sub fields, e.g., for pain assessment, deception detection, etc. (see e.g.,~\cite{li2020deep,pitaloka2017enhancing,miao2018chinese,DBLP:journals/mta/Gonzalez-Lozoya20}). Clearly, automated methods for ER have a great potential impact in a wide variety of fields.  

Perhaps the most studied and one of the most useful modalities for ER from an affective computing perspective is that comprising visual information, including still images and sequences.  In particular, facial expression recognition (FER) focuses on the analysis of facial imagery with  the aim of building predictive models to match faces with emotions. Traditional approaches to FER were based on \emph{standard} machine learning methods (e.g., support vector machines, neural networks etc.) feed with handcrafted features extracted from images, see e.g.,~\cite{670949,lucey2010extended}. 

Recently, methods based on deep learning methodologies, in particular those based on convolutional neural networks (CNNs), have been used  successfully to approach the FER task~\cite{li2020deep}. The benefits of these methods is that both, features and model, are learned simultaneously and there is no need of designing  handcrafted features.  Many solutions based on CNNs have been proposed recently, obtaining outstanding performance in a number of FER benchmarks. Despite its effectiveness, these methods may be overlooking important information that could be useful for improving the recognition process.  In particular, \emph{local patterns} that may arise in the context of FER and that are shared by subsets of subjects could be exploited to improve the recognition ability of the model. This is particularly relevant in FER where cultural, and ethnicity specific traits of faces make harder the task for \emph{global} CNN models; that is, generic models trained to learn common features for all of the subjects in the dataset under consideration. We hypothesize that enhancements to CNN models that take into account this local information may lead to improvements in performance, in particular, for minority class examples and samples that are difficult to classify with such \emph{global} models. 

In this paper we introduce an enhanced CNN based architecture equipped with multiple sets of radial basis function (RBF) units that aim at capturing local patterns. We call our proposed method a \emph{Multi Branch Deep RBF Network}. The proposed method comprises several branches of RBF units coupled with a standard CNN that acts as feature extractor, the outputs of the multi branch component are then concatenated and feed to a softmax layer that plays the role of classifier. The model is trainable end-to-end and inherits all of the benefits of CNNs for FER, with the additional advantage of exploiting local information. The proposed methodology is evaluated in a number of benchmark FER datasets and we show that the proposed enhancement outperforms considerably baselines that include a CNN without  any locality  component. In addition, the proposed method obtains state-of-the-art performance in some of the considered datasets, this is remarkable, given the \emph{simplicity} of the  considered backbone architecture (VGG-Face). 

The contributions of this paper are threefold: 
\begin{itemize}
    \item \textbf{The formulation of a variant of CNNs that aim at incorporating local information explicitly in the model.} As such, we are exploring the first steps of locally\footnote{The term \emph{local} in this context is with respect to samples in the dataset, and not to features like is the case in attention-based models.} weighted learning in the context of deep learning. To the best of our knowledge this is the first work adopting a locally weighted learning scheme in the task of FER.  Please note that there are very few efforts on locally weighted deep learning in general, see e.g.,~\cite{zadeh2018deep,papernot2018deep,vidnerova2018deep}.   

    \item \textbf{The introduction of  Multi Branch Deep RBF networks for Emotion Recognition.} This is an enhancement to CNNs that successfully exploits local information at the instances level, we show its effectiveness in the FER task.  % Each branch contains local learning modules, where each module consists of multiples RBF units to improve the model's predictive phase.  
    \item \textbf{An experimental evaluation} showing the 
    %We show that the 
    proposed methodology coupled with a standard architecture (VGG-Face) as backbone outperforms  baseline  models and achieves state-of-the-art performance that is comparable to more sophisticated and complex methodologies for some datasets.  More importantly we provide evidence that the competitive performance of the proposed model is due to the incorporation of local information. 
\end{itemize}

The findings and conclusions draw from this paper motivate further research on the study of locally weighted learning in the context FER and in general machine learning. 

The remainder of this paper is organized as follows. Section~\ref{sec:rw} reviews related work on FER and local learning. Next, Section~\ref{sec:method} introduces the proposed Multi Branch Deep RBF network model in detail. Then, Section~\ref{sec:results} presents an experimental evaluation of the proposed method in benchmark FER datasets. Finally, Section~\ref{sec:conclusions} outlines conclusions and future work directions. 

\section{Related work}
\label{sec:rw}
This section briefly reviews related work on FER and on the intersection of locally weighted learning with deep learning. 

\subsection{Emotion Recognition}
Automated ER is a task that has been studied for a while now, where the definition of emotion and a characterization of human emotions was inherited from the psychology field. Despite there are too many taxonomies and definitions, the most widely accepted categorization is that of Paul Ekman who defined six universal emotions: \emph{anger, disgust, fear, happiness, sadness and surprise}~\cite{ekman}; subsequently \emph{contempt} was also considered as a basic emotion. We adhere to this categorization in the remainder of the paper.   

The FER task was initially faced with a standard machine learning models feed with handcrafted features that aimed to capture discriminative facial characteristics. Common feature extractors comprise Histogram of oriented gradients (HoG)~\cite{gunawan2015face,chen2015facial,turan2018histogram}, Local Binary Patterns (LBP)~\cite{happy2012real,shan2009facial,zhao2007dynamic}, among others~\cite{ghimire2013geometric,ghimire2017recognition}.  While the considered classification models include 
Support Vector Machines (SVM)~\cite{suk2014real,georgescu2019local}, AdaBoost~\cite{ghimire2013geometric,chenadaboost} and Decision trees among others~\cite{wei2016real,dino2019facial}.  While competitive, most of the methods from the first wave relied on handcrafted features that not necessarily are representative or descriptive of the data \cite{maalej2011shape}.   
%\todo[author=HJ,inline]{@Fernanda: puedes anyadir referencias de otros trabajos que usen estos u otros descriptores y clasificadores?}

The advances in deep learning have motivated a second waive of methodologies that rely on deep learning~\cite{li2020deep}. Contrary to the traditional approaches, these methods simultaneously learn the representation for the input  and the predictive model. In this way, features are derived entirely form data and these are tied to the classification model being learned. The most commonly used deep learning architectures that have been used for ER are Convolutional Neural Networks (CNNs), see e.g.,~\cite{chen2019facial,shima2018image,videla2020facial,ravi2020face,shao2021fcnn,mollahosseini2016going,li2017reliable}. 
%\todo[author=HJ,inline]{@Fernanda: Incluir todas las referencias que citamos y que usan CNNs aqu\'i}
These models take as input raw images and learn multiple layers of convolutional filters that are applied to the inputs of the layer and their outputs feed to the next one. These models are coupled with other types of layers including dense and softmax layers to learn the predictive part of the model. Other popular architectures comprise Residual Networks~\cite{sepas2019deep,wang2020suppressing,farzanehfacial,wang2020region,li2019separate} and sequential models (e.g., LSTMs)\cite{zhang2020facial}.
%\todo[author=HJ,inline]{Incluir aqu\'i referencias que se basen en ResNets, LSTMs o incluso otro tipo de redes}
Additionally, these methods are able to incorporate additional mechanisms and features into the learning process making them quite effective and self contained, for instance, architectures with attention mechanisms~\cite{li2020attention,farzanehfacial,wang2020region,minaee2019deep}, \emph{ad hoc} loss functions~\cite{sepas2019deep} and other complex procedures ~\cite{liang2020fine,zeng2018facial,zhao2016peak}.
%\todo[author=HJ,inline]{@Fernanda: aqu\'i se tiene que listar todas las referencias que citamos y que usen mecanismos especiales}
The author is referred to~\cite{li2020deep} for a comprehensive survey on FER with deep learning based methodologies.

\subsection{Deep Learning models with local learning}

Despite the effectiveness of  deep learning based solutions, there are still open questions that deserve attention from the community and that could have a great impact into the field. One of these  questions has to do with the lack of specific mechanisms in CNNs for taking into account local information at the \emph{instance level}. Local information has proven to be very helpful in classical models within  machine learning. Consider for instance locally weighted regression~\cite{DBLP:books/daglib/0087929}, where the incorporation of samples \emph{close} to a query sample are used to approximate a regression function locally. This feature enables a linear model (e.g., least squares regression) to approximate non linear decision surfaces. There are many other cases where local information has proven to be very useful including support vector machines~\cite{10.5555/3104482.3104606}, learning vector quantization~\cite{DBLP:journals/nca/NovaE14},  decision trees and even there is a variant of neural network that implements locally weighted learning: the radial basis function network~\cite{DBLP:books/daglib/0087929}. The next section reviews related work on some efforts on locally weighted learning for deep learning. 
%\todo[author=HJ,inline]{@Fernanda: aqu\'i anyade por favor referencias de modelos tradicionales con info loca, como DTs, NNs, etc.}

In the context of deep learning, local learning has been scarcely studied in the context of  CNNs and other Deep Neural Networks (DNNs). Zadeh et al. introduced a 
deep RBF model that aimed to make robust predictions against adversarial attacks~\cite{zadeh2018deep}. The model is formed by a CNN architecture tied with RBF units in the output layer (these were used to perform classification). A loss function tailored to be robust against adversarial attacks was proposed. Vinderov\'a et al. presented a method where a DNN and RBF networks are concatenated  to classify adverse examples correctly~\cite{vidnerova2018deep}. Although these efforts combine deep learning with locally weighted learning the aim is not to improve the predictive performance of the model in general, but for specific \emph{adversarial} scenarios. Moreover in~\cite{vidnerova2018deep} the DNN and the RBF network are trained separately, which results in the combination of two independent models.

There are other efforts from the community that aim at taking advantage of other locally weighted learning mechanisms in the context of deep learning. For instance, Bahri et al. proposed a Deep k-Nearest Neighbor (DkNN) model for detecting noisy examples~\cite{bahri2020}. Earlier, Papernot et al. introduced a DkNN model that estimated neighbors across layers aiming to have additional information for the predictive model (interpretability)~\cite{papernot2018deep}. Like the methods in~\cite{zadeh2018deep,vidnerova2018deep}, the DkNN methods target adversarial or noisy examples. Other efforts like that that of Yu et al. have applied the principle of local learning (a model is trained only on the most relevant data for a given input) for very specific scenarios~\cite{yu2017fine}. 

\subsection{Discussion}

The FER task has been approached for a while and the most effective solutions are those based on deep learning methodologies. These methods have the appealing features that they can learn simultaneously features and predictive model. While these models have obtained outstanding performance, there are several questions around these models that deserve to be explored. In this paper we aim to explore the benefits of locally weighted learning (LWL) into deep learning models for approaching the FER task.  LWL has been scarcely studied in the context of deep learning. There are few efforts in this direction and all of them target very specific scenarios, for instance, classification with adversarial examples and interpretability. We argue that LWL could be beneficial for FER because there are samples that in order to be correctly classified, the model should build a \emph{sub-classifier} that considers only samples similar to the  query point. Intuitively, consider ER datasets in which a minority group is underrepresented, and dominated by another group (see Section~\ref{sec:resultsbaselines} for an example). \emph{Global} models are prone to fail to correctly classify such instances and this type of issues could be alleviated with LWL. For these reasons we propose in this paper a LWL deep learning model  that does not target specific scenarios (like adversarial samples). We show the potential of this model in an experimental evaluation. %The proposed model is introduced in the next section. 

\section{Multi-Branch Deep Radial Basis Function Networks}
\label{sec:method}

The working hypothesis of this work is that the incorporation of local information at the instance level into the learning process of CNN based models improves the recognition performance of the enhanced model for the FER task. The intuition behind this hypothesis is that in FER there may exist groups of instances that share similarities to each other in one or more aspects (e.g., in terms of ethnicity or age). Therefore, when classifying a query sample, a prediction that is build by taking into account information of similar instances should improve the recognition performance.

\begin{figure*}[h!]
    \centering
    \includegraphics[width=0.8\linewidth]{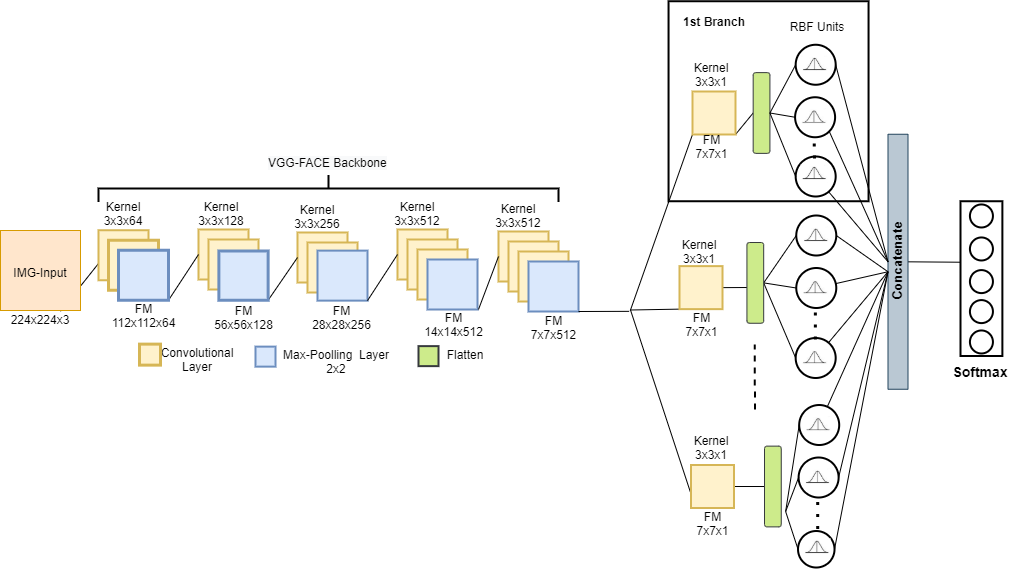}
    \caption{Diagram of the proposed Multi-branch Deep Radial Basis Function Network architecture. A backbone architecture (VGG-Face in this case) is enhanced with multiple branches of RBF units which are then connected to a dense softmax layer. }
    \label{fig:DRBF}
\end{figure*}

We introduce in this section a model that implements such an idea, the so called \emph{Multi-Branch Deep RBF} Network model. In a nutshell, the model uses a CNN as backbone (e.g., VGG-face) and it is enhanced with a new layer formed by multiple branches of  RBF units. Such RBF layer receives as input feature maps from the preceding layers of the CNN and its outputs are concatenated and  connected to a softmax layer that  makes predictions over the considered classes. This enhancement allows a CNN to implicitly incorporate local information that can have a positive impact  in  recognition performance. A graphical diagram of the proposed model is depicted in Figure~\ref{fig:DRBF}, the remainder of this section describes the proposed model in detail. 

\subsection{Radial basis function networks}
An RBF unit is type of neuron that is associated to a center and a radius, it can be considered as a prototype in the input space whose position is updated (learned) from data. These units are commonly used in RBF networks and LVQ-based models. In the case of the former models, a set of units defines a layer, and commonly there is an RBF unit per class associated to the problem at hand (i.e., RBF units are often used instead of softmax ones for the predictive part of the model).  The output $h_i $ of an RBF unit $i$ given input $\mathbf{x}_j$ is computed as follows:
\begin{equation}\label{eq:rbfunit}
h_i = \phi_i(\mathbf{x}_j) = e^{-  \left( {\left \| \mathbf{x}_j - \mathbf{\mu}_i \right \|}^2  \times (2 \sigma{i}^2)^{-1}  \right)} 
\end{equation}

where $\mathbf{x}_j \in \mathbb{R}^d$ is a $d-$dimensional feature vector, and $\mathbf{\mu_i} \in \mathbb{R}^d$, $\sigma_i \in \mathbb{R}$ are the center and radius of RBF unit $i$. 

\subsection{Multi-branch RBF layer}
As previously mentioned the proposed model extends a backbone CNN with RBF units arranged into branches as shown in Figure~\ref{fig:DRBF}. We relied on VGG-Face~\cite{Parkhi15} as backbone because is a well known and \emph{generic} enough CNN for facial analysis that has proven to be very helpful in FER and related tasks when used as pre-trained model. \textcolor{black}{One should note that VGG-Face is not a state-of-the-art methodology as those that are being currently proposed in the context of FER (see Section~\ref{sec:rw}). Our decision for relying in a generic model lies in that we wanted to prove the proposed extension could lead to improvements with a standard model. Relying on more complex or elaborated models would make it more complicated to assess the actual improvement due to our local modules. } 

VGG-Face is an architecture formed by  a series of convolutional layers followed by fully connected layers that are in turn followed by a softmax layer in charge of the classification process~\cite{Parkhi15}. We modify the last few layers of backbone architecture as follows. We dropped all of the fully connected layers, and instead connected multiple branches of RBF units (see Expression (\ref{eq:rbfunit})) to the output of the last convolutional layer. Such convolutional layer (see Figure~\ref{fig:DRBF}) returns as output 512 feature maps of dimensionality $7 \times 7$. We take the activation of these maps as inputs to the multiple RBF branches. 

The motivation behind having multiple branches of RBFs is that if one would have a single RBF, the input would be very high dimensional (i.e., $7 \times 7 \times 512= 25088$) and potentially useful local information would get lost or would be very difficult to process. In contrast, having multiple branches of RBFs each taking as input a low dimensional vector could lead to exploit local information easily. In fact, it would be expected that each branch could capture a local pattern different from the rest (see Section~\ref{sec:viz}). Therefore, we propose to process the outputs of the last convolution layer in such a way that each branch takes an input of manageable size. Specifically, we process the last convolutional layer with a set of filters that yield a $7 \times 7$ output and we use as many of these filters as branches are considered in the model (see Figure~\ref{fig:DRBF}). These outputs are flattened and feed to the RBF branches in a fixed order.

In the proposed model, the pre-trained convolutional layers of the VGG-Face model are used~\cite{Parkhi15}.  The RBF centers and their corresponding radius are initialized randomly. The  model is then trained end-to-end using backpropagation and stochastic gradient descent (more details in Section~\ref{sec:results}) using the FER dataset at hand. We performed experiments on different ways of freezing and updating weights for the whole architecture, we observed there was no significant difference in performance when updating or not the (convolutional) weights inherited from the backbone architecture, therefore we decided to froze the weights of VGG-Face and learn only the remainder of the parameters.  
One should note that even when the displayed in Figure~\ref{fig:DRBF} shows a branch per feature map, a reduced number of branches could be used as well, see Section~\ref{sec:results}. 

\begin{figure}[h!]
    \centering
    \begin{tabular}{ccc}
    \subfloat[Module with RBF layer.\label{Fig:ReductionA}]{\includegraphics[width=0.4\linewidth]{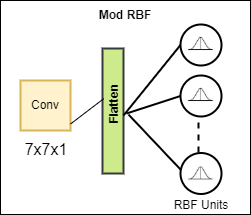}}
    &&
    \subfloat[Module with dense layer.\label{Fig:ReductionB}]{\includegraphics[width=0.4\linewidth]{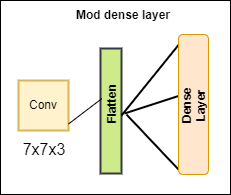}}
    \end{tabular}
    \caption{Illustration of units used in the proposed model (left) and the CNN variant, Multi-branch CNN (right). Each module has a convolutional filter followed by a flatten layer. The difference of each module is the layer used after the flatten layer:  a) uses RBF layer while b) uses dense layer with a ReLU activation function. }
    \label{fig:Branch_Mod}
\end{figure}

In the next section we present an experimental evaluation that shows the proposed model outperforms strong baseline models. %s including VGG-Face trained from scratch and fine tuned. 
In particular, we compare the proposed model to a reference model that replaces the RBF branches by dense layers, this is illustrated in Figure~\ref{fig:Branch_Mod}. This is important to mention as this comparison will allow us to determine the actual benefits of having local information instead of fully connected units as it is standard in CNN models. 

\subsection{Discussion}
We just introduced the  \emph{Multi-Branch Deep RBF} Network model: an enhancement to CNNs in which a layer formed by several branches of RBF units is added before the softmax classification layer. The main novelty of this proposal is the adoption of multiple RBF sub networks that allow the model to deal with the high dimensionality of the input space, likewise, having multiple branches allow the model to capture specific local information in each of these. Compared to alternative solutions from deep RBF networks, which use a single branch of RBF units, in our model the outputs of multiple branches are feed to a softmax layer that makes predictions, whereas in reference work (see e.g.,~\cite{zadeh2018deep,vidnerova2018deep})   RBF units are used to make the predictions directly. Additionally, one should note that the focus of previous work has been on using local information for under covering adversarial attacks (see, e.g., ~\cite{zadeh2018deep,vidnerova2018deep,papernot2018deep}), while in this paper our goal is to improve the overall classification process. Finally, to the best of our knowledge this is the first effort on trying to incorporate local information at the instance level into the task of FER. As shown in the next section the proposed enhancement improves considerably the performance of reference models and performs favorably with state of the art solutions that are based on much elaborated mechanisms and models (e.g., attention based models).

\section{Experimental evaluation}
\label{sec:results}
This section presents an experimental evaluation of the proposed model in the FER task, the goal is to show the competitiveness and benefits of the  \emph{Multi-Branch Deep RBF} Network model when compared to reference models and to state-of-the-art solutions.  We first introduce the datasets and experimental settings; then, we present an ablation study that analyzes the performance of our model under different parameter settings; next we show some visualizations that aim at highlighting the benefits of our model; finally, we compare the performance of our model to reference and state-of-the-art methods and conclude with a discussion. 

\subsection{Experimental settings}
\begin{figure}[h!tb]
    \centering
    \includegraphics[width=1\linewidth]{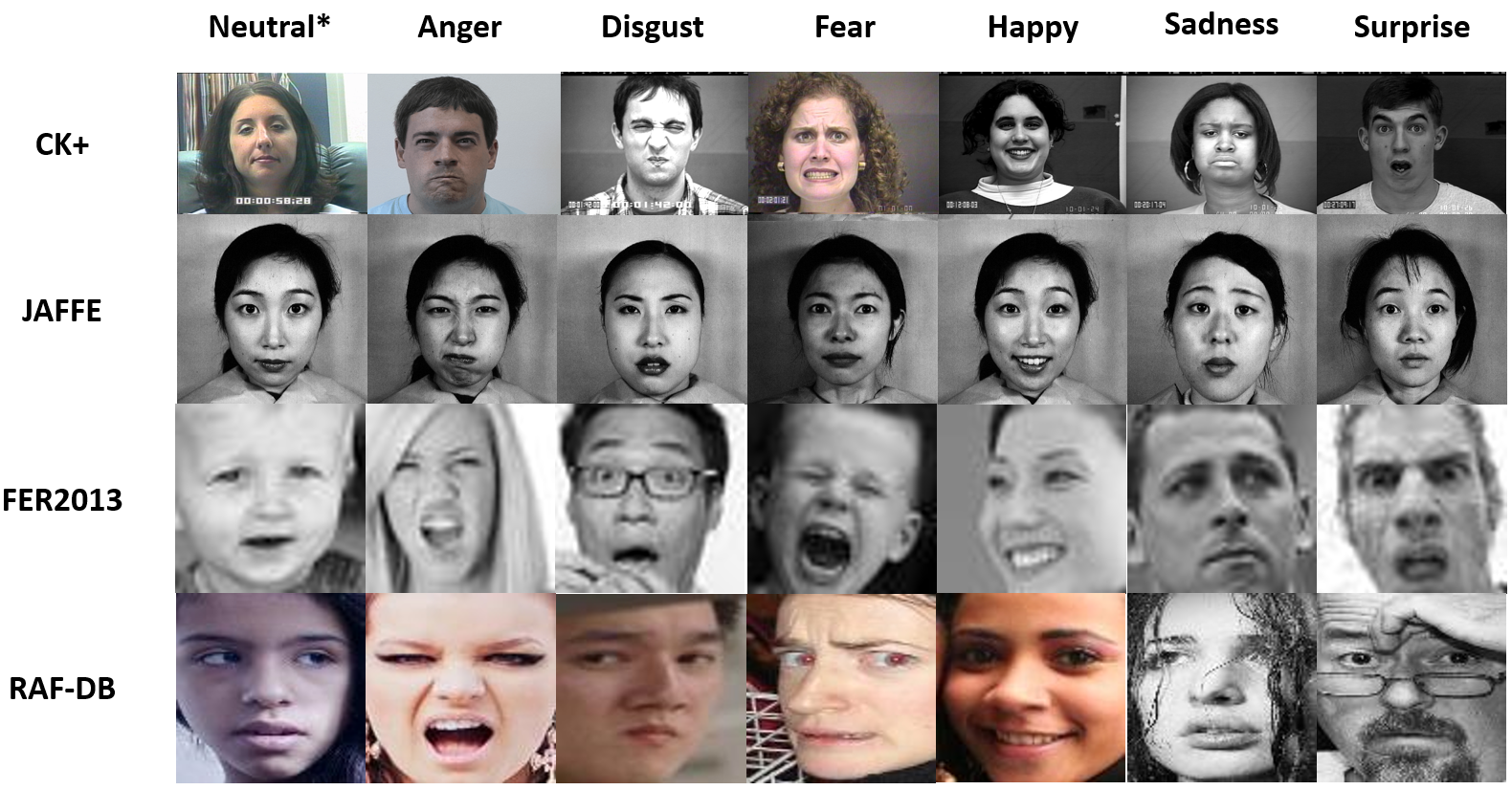}
    \caption{Sample images associated to different emotions for the CK+, JAFFE, FER2013 and RAF-DB datasets. Please note that for the CK+ dataset we are showing an image of the \emph{Contempt} emotion instead of the \emph{Neutral} one, as it is the only dataset without the latter emotion. }
    \label{fig:datasets1}
\end{figure}

\begin{figure}[h!tb]
    \centering
    \includegraphics[width=0.95\linewidth]{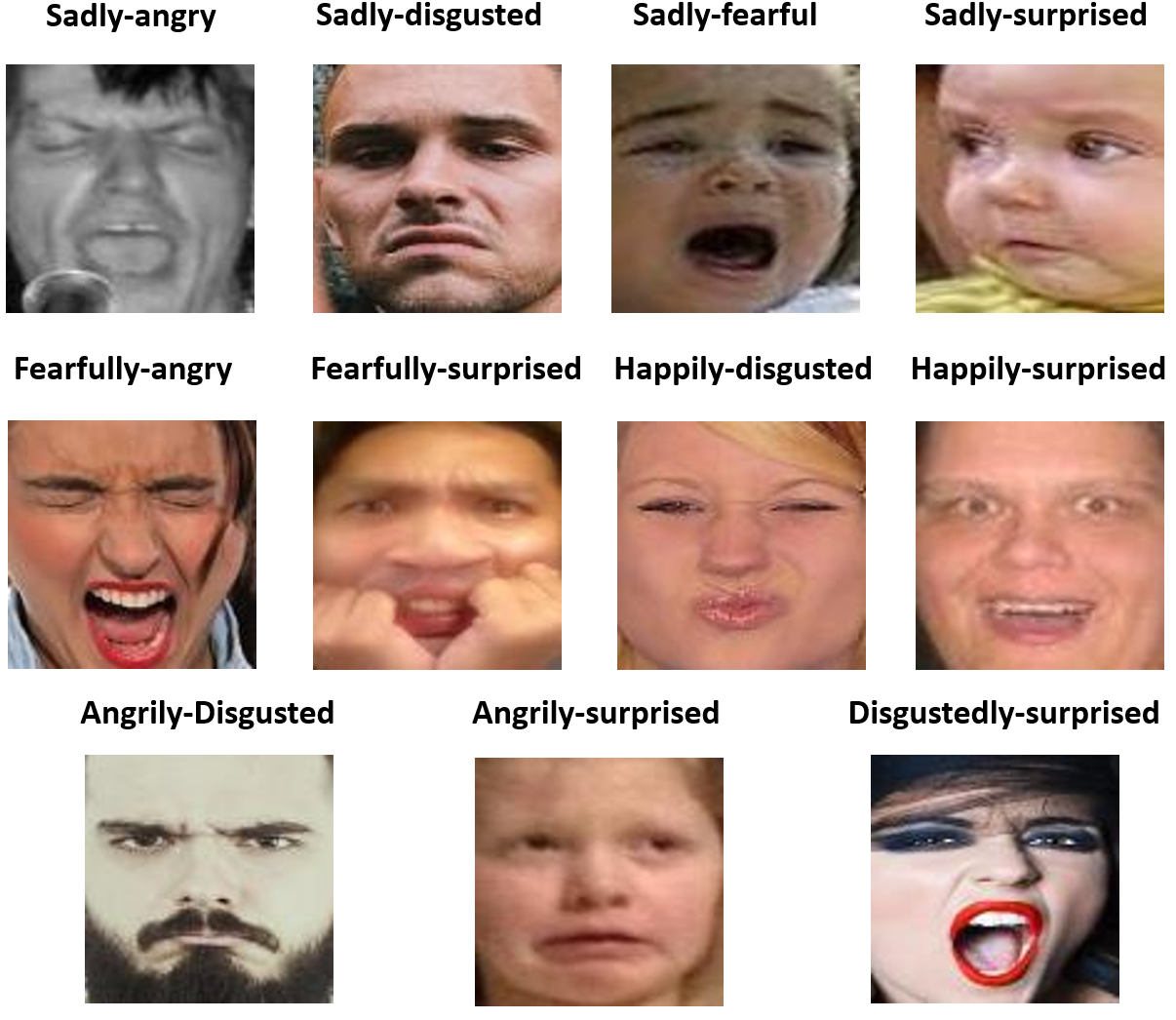}
    \caption{Sample images associated to different emotions considered in the RAF-DB Compound dataset. }
    \label{fig:datasets2}
\end{figure}

For the experimental comparison we used the following benchmark datasets that have been widely used in the literature (see~\cite{li2020deep}): Real-world Affective Faces Database (RAF-DB), RAF-DB Compound~\cite{li2017reliable,li2019reliable}, the Extended Cohn-Kanade Dataset (CK+)~\cite{lucey2010extended}, the Japanese Female Facial Expression (JAFFE) Dataset~\cite{670949,lyons_michael_1998_3451524}, and FER 2013~\cite{goodfellow2013challenges}. Additionally we performed experiments in a challenging dataset combining both  CK+ and JAFFE datasets. Samples from the considered datasets are shown in Figures~\ref{fig:datasets1} and~\ref{fig:datasets2} and some statistics are presented in Table~\ref{tab:datasets}.

The considered datasets comprise a diversity in terms of the number of samples, background/recording conditions, and complexity. Where one should distinguish datasets under the \emph{standard} ER setting from datasets of greater difficulty. \emph{Standard} datasets including CK+, JAFFE, FER2013 and RAF-DB comprise basic emotions\footnote{Please note hat CK+ includes the \emph{Contempt} emotion instead of the \emph{neutral} one, see Figure~\ref{fig:datasets1}.} and images coming from the same distribution.  The challenging datasets are CK+-JAFFE and RAF-DB Compound, the former made up by merging images of the CK+ and JAFFE datasets, and the latter considering a fine-grained classification of emotions, see Figure~\ref{fig:datasets2} . 
 
The intuition behind experimenting with the merged CK+-JAFFE dataset lies in that we wanted to assess the performance of our model when there are clear differences across samples from the same category.  

% \begin{figure}[htbp]
%     \centering
%     \begin{tabular}{c}
%     \begin{tabular}{ccc}
%     \subfloat[CK+\label{Fig:ck+}]{\includegraphics[width=0.3\linewidth]{Fig_CK+.pdf}}&
    
%     \subfloat[JAFFE\label{Fig:jaffe}]{\includegraphics[width=0.3\linewidth]{Fig_JAFFE.pdf}}&
    
%     \subfloat[RAF-DB\label{Fig:raf}]{\includegraphics[width=0.3\linewidth]{Fig_RAF-DB.pdf}} \\
%     \end{tabular}\\
%     \begin{tabular}{cc}
%      \subfloat[FER 2013\label{Fig:fer2013}]{\includegraphics[width=0.4\linewidth]{Fig_FER 2013.pdf}}&
     
%     \subfloat[RAF-DB Compound\label{Fig:rafcomp}]{\includegraphics[width=0.4\linewidth]{Fig_RAF-DB-Compound.pdf}}\\ 
%     \end{tabular}
%     \end{tabular}
%     \caption{Sample images associated to different emotions for the datasets considered for experimentation.}
%     \label{fig:datasets}
% \end{figure}

\begin{table}[h]
    \centering 
    \caption{Statistics of the datasets considered for experimentation. The number of classes (E) and training, validation, and test samples are shown. Please note that for CK+-JAFFE the number of classes is 8 as we include the \emph{Contempt} emotion that is only present in CK+.
    }
    \begin{tabular}{|c|c|c|c|c|}
    \hline
    \textbf{Dataset}   & \textbf{\#Tr.} & \textbf{\#Val.} & \textbf{\#Test} & \textbf{\# E} \\ \hline
    \textbf{CK+}~\cite{lucey2010extended}       & 877                         & 94                            & 123                        & 7                \\ 
    \textbf{JAFFE}~\cite{lyons_michael_1998_3451524}    & 143                         & 35                            & 35                         & 7                \\ 
    \textbf{CK+JAFFE}  & 1,020                       & 129                           & 158                        & 8                \\ 
    \textbf{FER 2013}~\cite{goodfellow2013challenges}  & 28,709                      & 3,589                           &  3,589                       & 7                \\ 
    \textbf{RAF-DB}~\cite{li2019reliable}    & 12,271             & 3,068                            & 3,068                         & 7                \\ 
    \textbf{RAF-DB Compound}~\cite{li2017reliable}    & 3,162                         & 792                            & 792                         & 11                \\ \hline
    
    \end{tabular}
        \label{tab:datasets}
\end{table}
On the other hand, the RAF-DB Compound is challenging because %all but the RAF-DB Compound dataset are labeled with the seven basic emotions: \emph{anger, disgust, fear, happiness, sadness, surprise and contempt}. While the remaining dataset
it considers compound emotion categories (e.g., \emph{Fearfully-surprised, sadly-angry, happily-disgusted}, etc), 11 categories are considered, see Figure~\ref{fig:datasets2}. The idea of considering this dataset is to show the benefits of incorporating local information into the recognition process for approaching a fine grained FER task.  It is expected that the proposed model is more advantageous in the two considered challenging datasets. 

For all of the datasets we used the top-1 accuracy on the test set as the evaluation measure of performance. This is in agreement with previous work using the same datasets. The same partitions for training and testing were used in datasets where these were available (RAF-DB, RAF-DB Compound and FER2013) and random splits of 80\% for training and validation and 20\% for testing were used CK+ and JAFFE. For the latter datasets multiple partitions were generated and their results averaged in each experiment. %For datasets were 

The model was trained  using the Adam~\cite{kingma2014adam} optimizer with a batch size of 32 during 100 epochs. The performance in validation was used to monitor convergence of the model. We determined  the value of $\sigma$ experimentally as $\sigma=0.0528$.  The model was trained in a laptop with a Nvidia GTX card 2080 with 8Gb of VRAM, and a processor I7 6700K with 32 Gb of RAM.
\begin{figure*}[h!tb]
    \centering
    \begin{tabular}{c}
    \begin{tabular}{cc}
    \subfloat[CK+\label{Fig:Resultados_ck+}]{\includegraphics[width=0.5\linewidth]{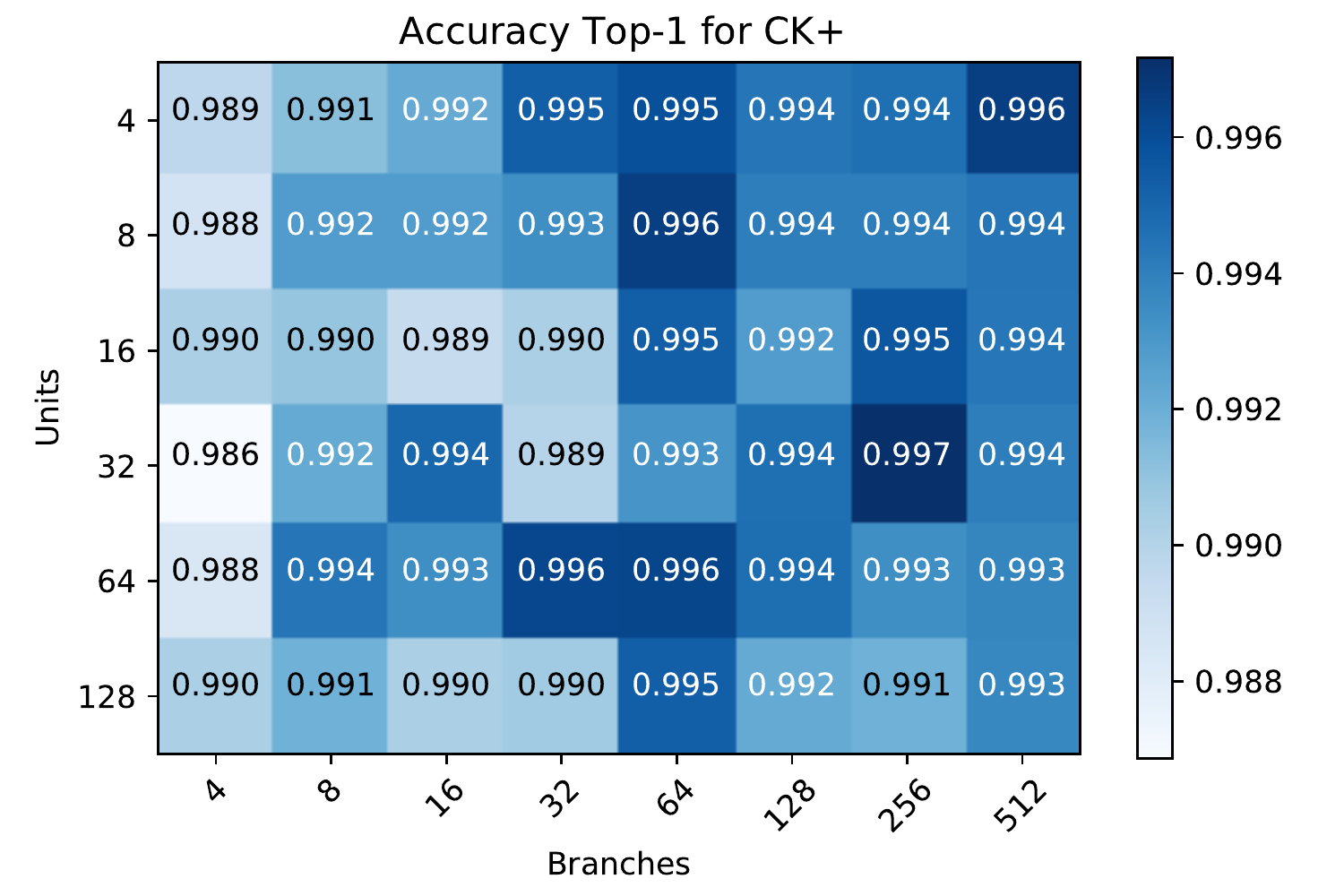}}&
    
    \subfloat[JAFFE\label{Fig:Resultados_jaffe}]{\includegraphics[width=0.5\linewidth]{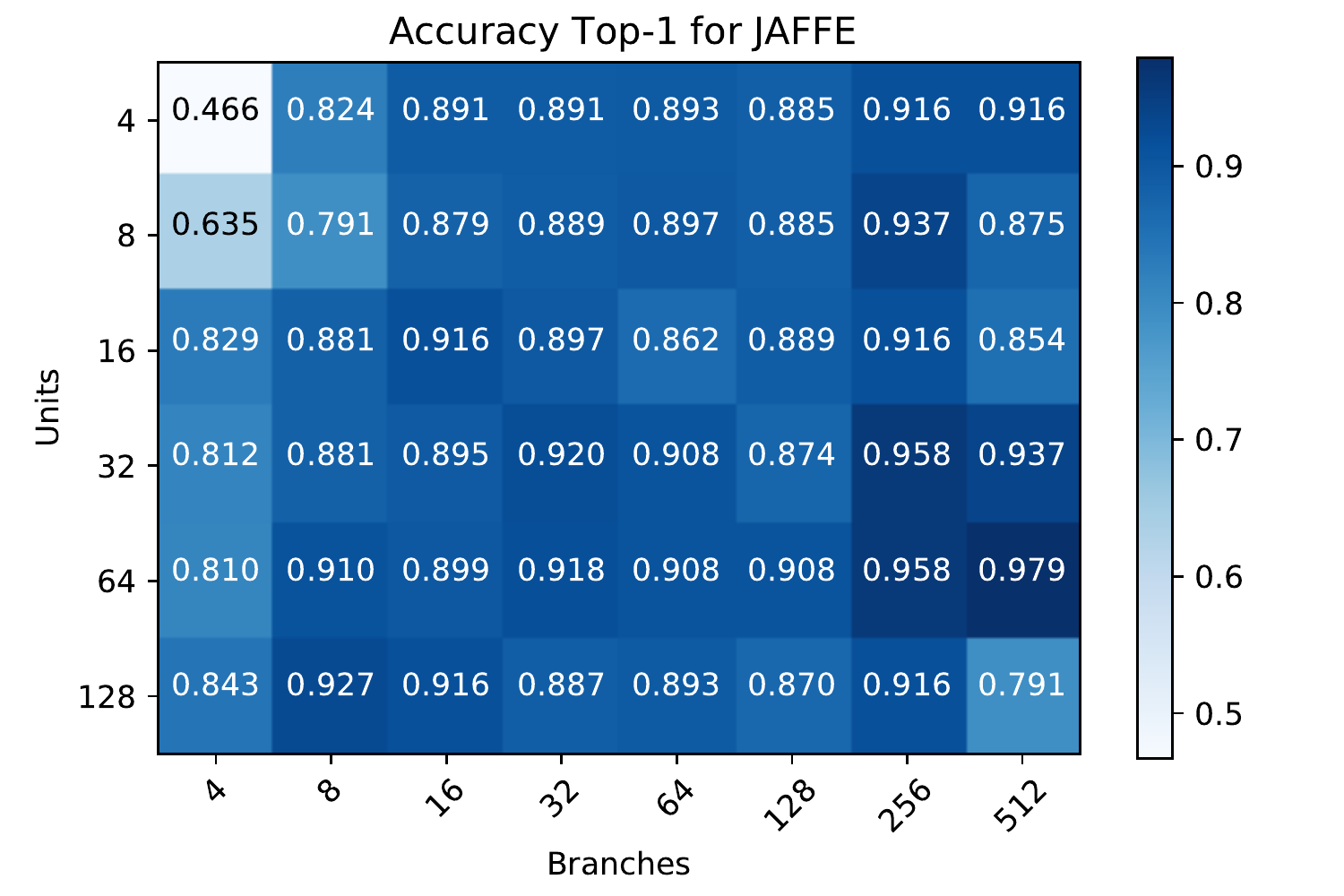}}

    \end{tabular}\\
    \begin{tabular}{cc}
    
    \subfloat[CK+JAFFE\label{Fig:Resultados_CK+JAFFE}]{\includegraphics[width=0.5\linewidth]{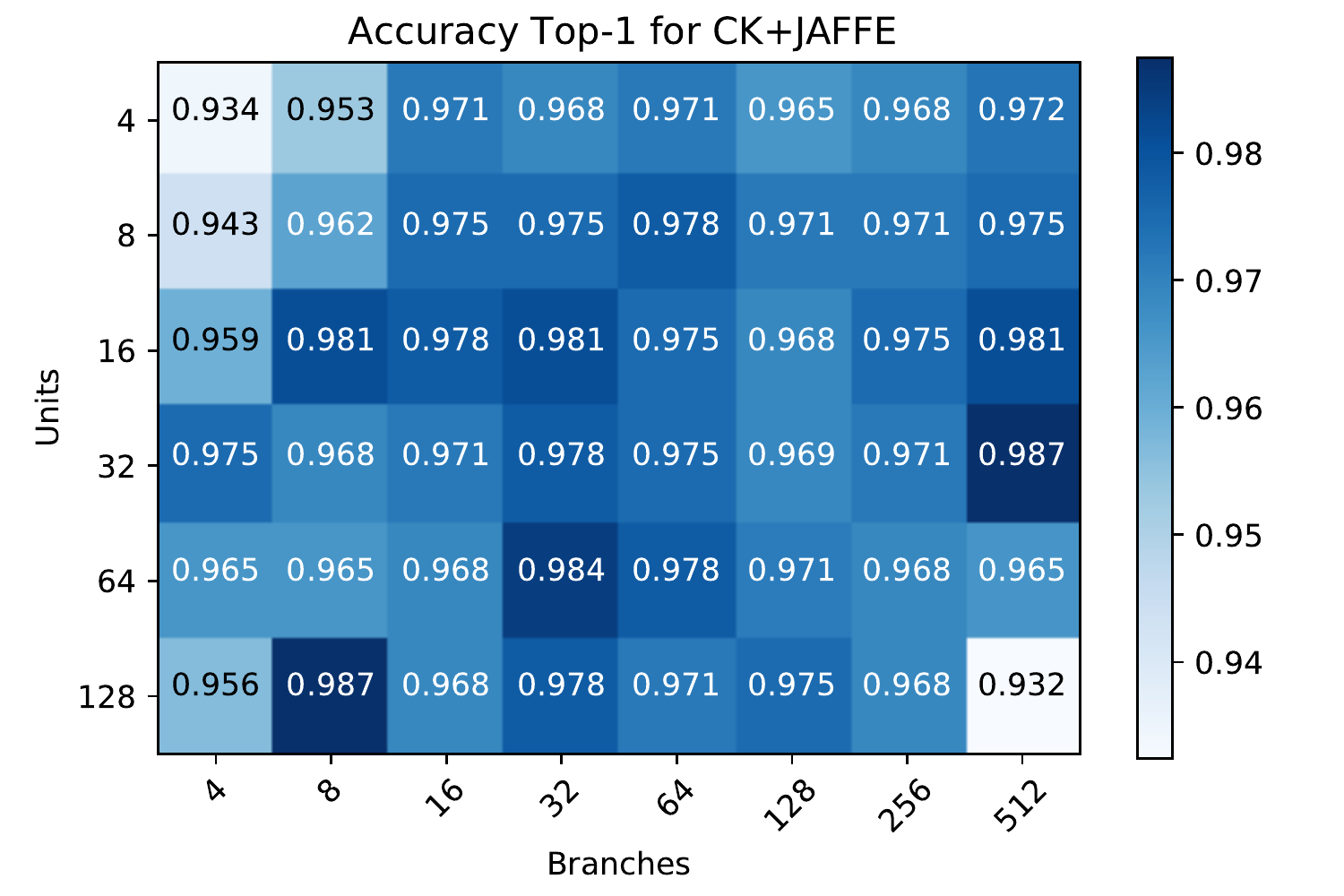}} &
    
     \subfloat[FER 2013\label{Fig:Resultados_FER2013}]{\includegraphics[width=0.5\linewidth]{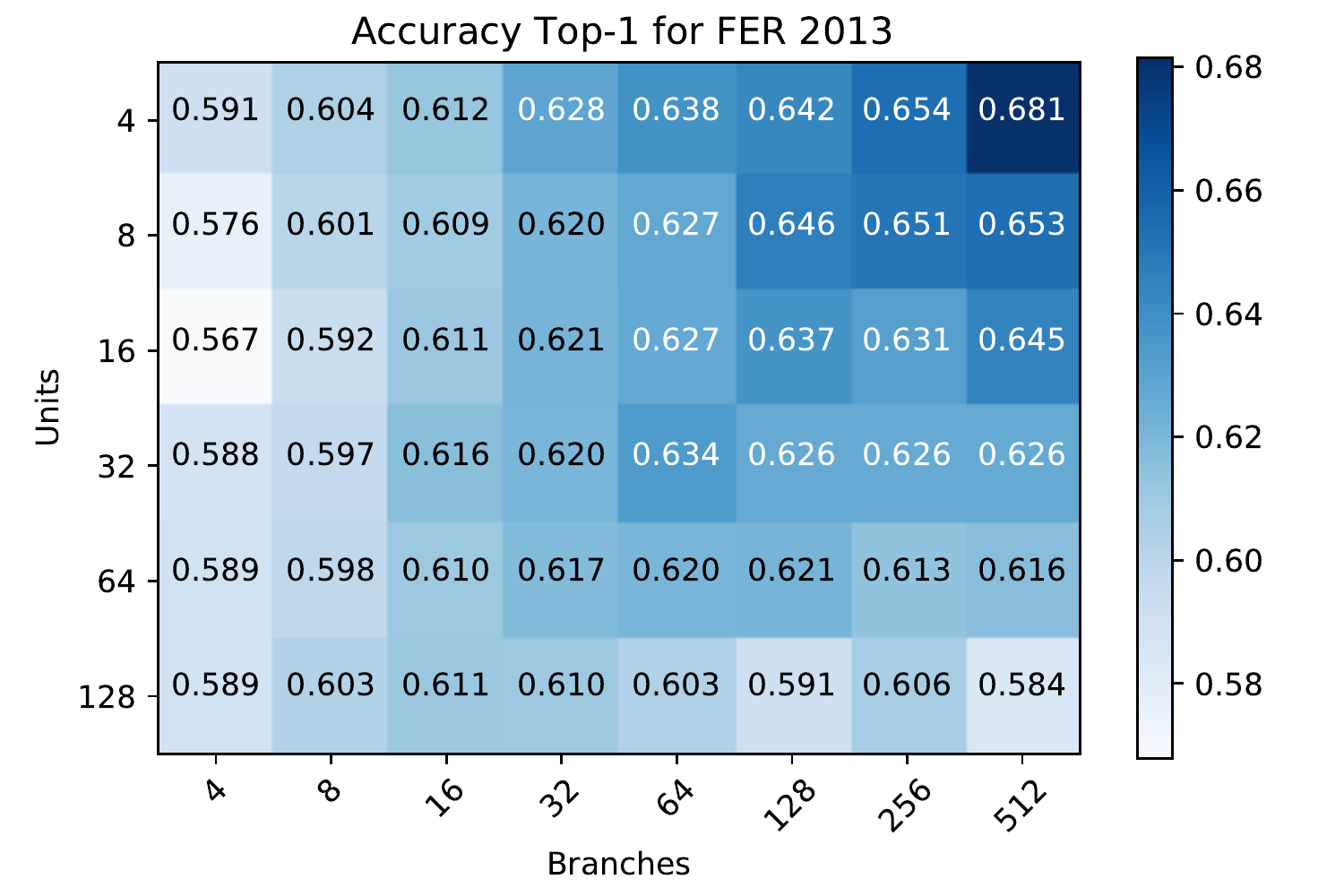}}
    \end{tabular}\\
     
    \begin{tabular}{cc}
     
     \subfloat[RAF-DB\label{Fig:Resultados_RAF}]{\includegraphics[width=0.5\linewidth]{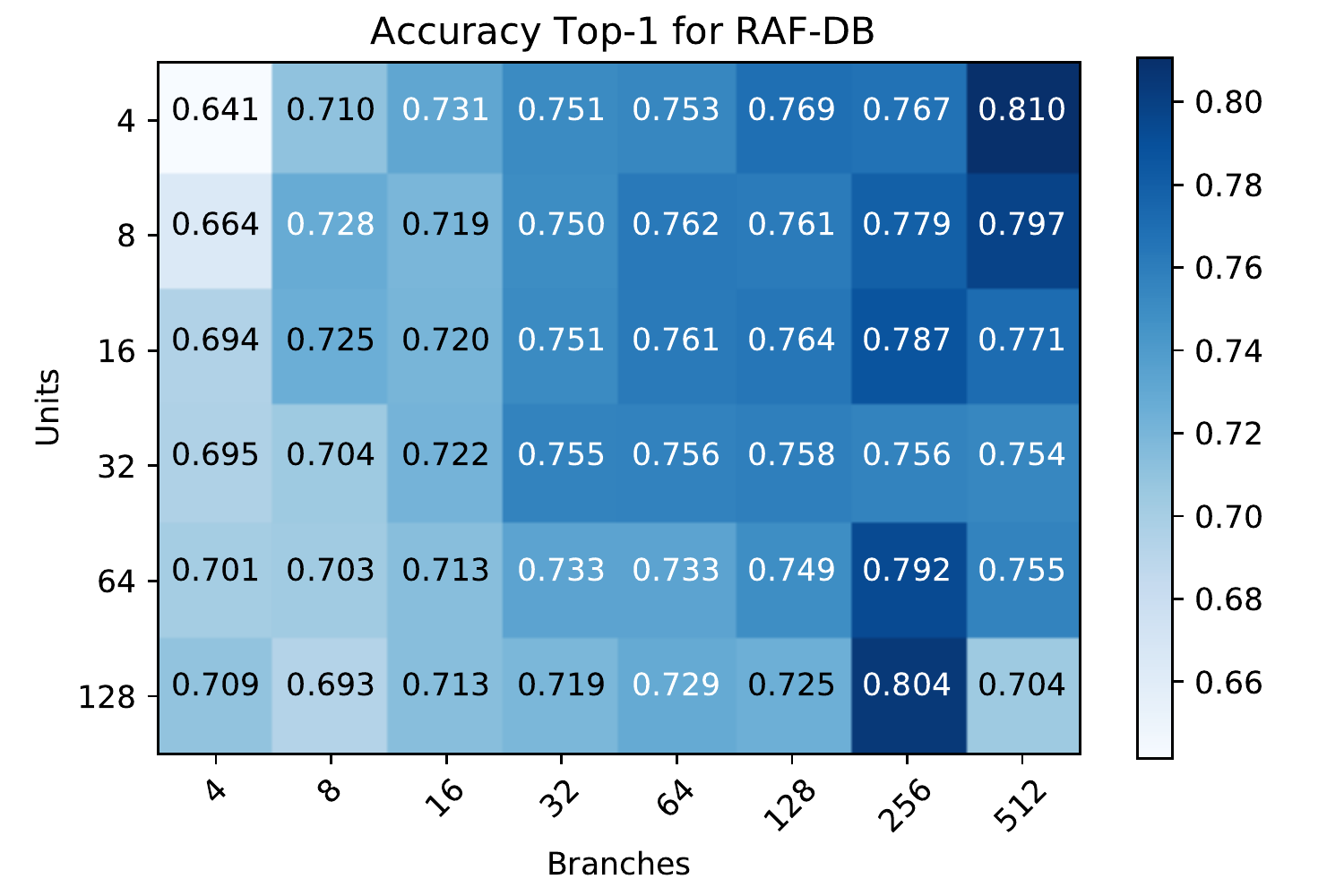}}
     &
     \subfloat[RAF-DB Compound\label{Fig:Resultados_RAF_Compun}]{\includegraphics[width=0.5\linewidth]{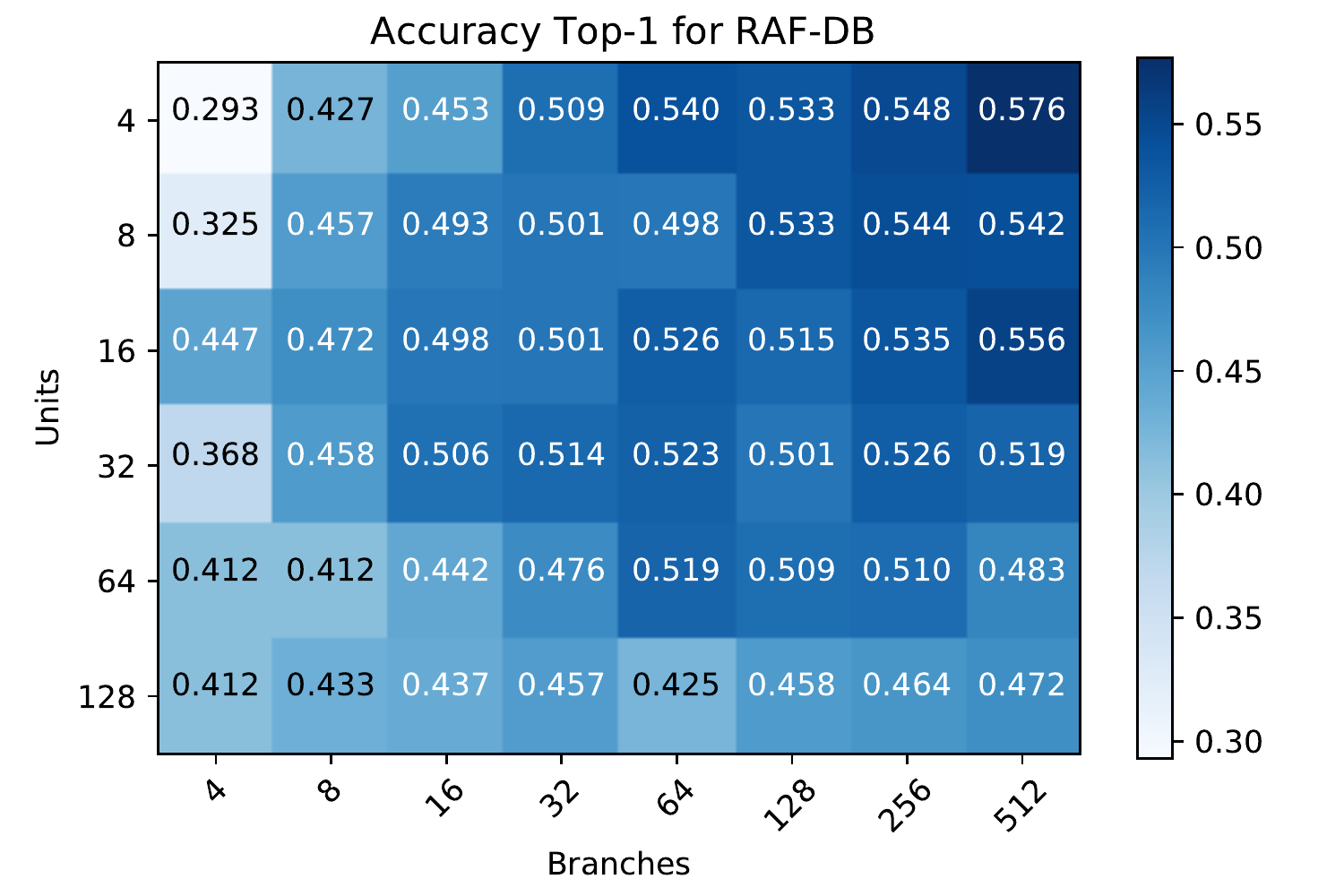}}
    \\ 
    \end{tabular}
    \end{tabular}
    \caption{FER recognition performance when varying the number of branches and RBF units per branch in the proposed model.}
    \label{Fig:ResultsHeatMap}
\end{figure*}
\subsection{Ablation study}
In this section we evaluate the performance of the proposed model when varying the number of branches and units. We present in Figure~\ref{Fig:ResultsHeatMap} the results of this evaluation for the six considered datasets. Results are shown as heat maps (the darker the better), the number of branches is specified in the $x$ axis and the number of unites is shown in  the $y$ axis.

As it can be seen from this figure, mixed results are obtained for the different datasets. Being the CK+ dataset the \emph{easiest} and RAF-DB Compound the \emph{toughest} in terms of recognition performance. The difference between the lowest and highest performance achieved for every dataset makes clear that it is necessary to adequately tune both of these parameters (e.g., compare the lowest and highest performance in Figures~\ref{Fig:ResultsHeatMap} (b) and (f)). 

Although no general conclusion can be drawn on the values of parameters, a pattern that seems to be present in all of the datasets is that a larger the number of branches seems to result in better performance of the model. Also, it seems that a small number of RBF units combined with large number of branches is a somewhat robust combination of parameters.  In general, the obtained performance in most datasets is competitive with the state-of-the-art (see Section~\ref{sec:sota}). For the experiments reported in the next sections the best configuration of parameters for each dataset was used. 

\subsection{Visualization of centers}
\label{sec:viz}
We now present visualizations of learned RBF centers for two configurations of parameters of the proposed model for the CK+-JAFFE  dataset. We chose this particular dataset because it is one formed by instances from two different datasets and we expect the local information to be particularly helpful (see Section~\ref{sec:sota}). Also, one should note that performance for this dataset did not vary too much for the different choices of parameters as shown in Figure~\ref{Fig:ResultsHeatMap} (c).

\begin{figure}[h!tb]
    \centering
    \includegraphics[width=0.95\linewidth]{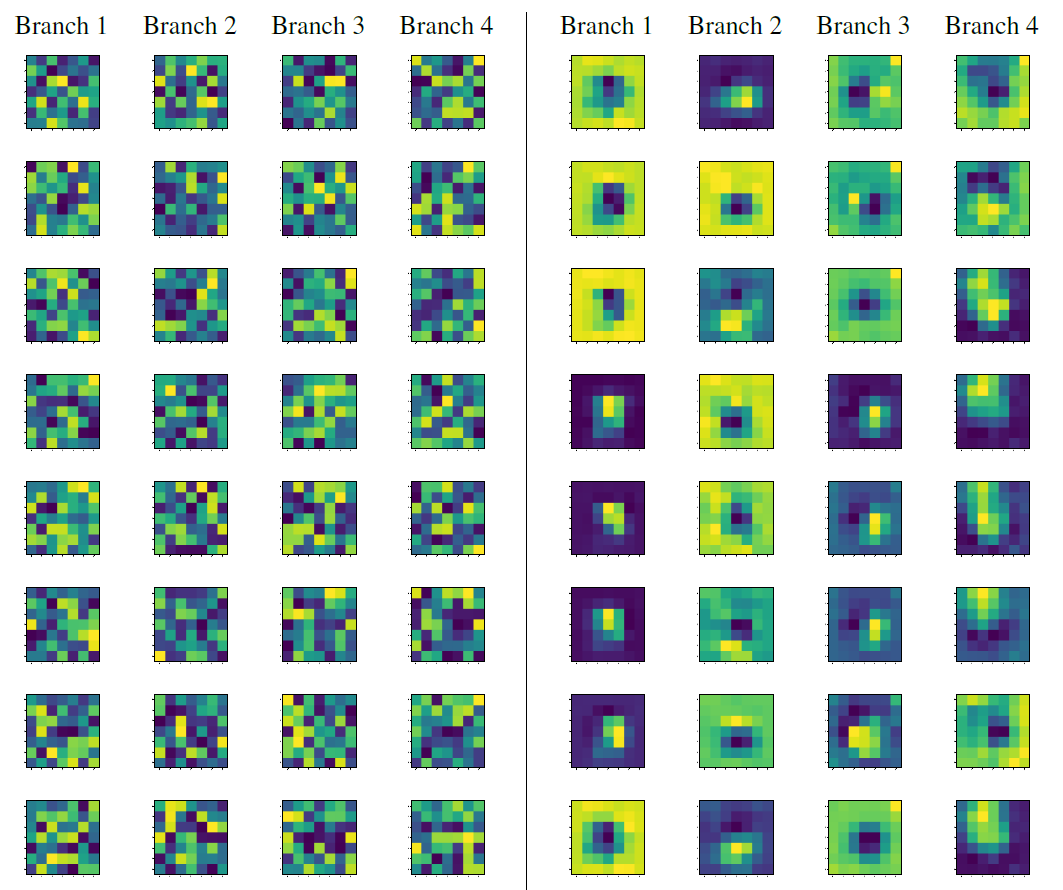}
    \caption{Visualization of the centers of the RBF units for a model with 4 branches and 8 RBF units. The left plot shows the initialized centers and the right one shows the learned centers after convergence.   }
    \label{fig:4ramas}
\end{figure}

Figure~\ref{fig:4ramas} shows the centers for a configuration with 4 branches and 8 RBF units per branch, the reported performance for this configuration was 0.943. From this figure it can be seen that centers across the branches are very different to each other. \textbf{\emph{Corroborating the hypothesis that different centers are modeling different aspects of the input feature maps}}. It is only for branch 1 that there seem to be similarities among centers (column 1, rows 4-7 of the left plot). In general, it seems that the relevant information is located near the center of the image (blue values in the center, yellow for the background), which make sense given the approached task is ER. However, there are a few centers that are also giving large weights to the region surrounding the face (blue background). 
\begin{figure}[h!tb] 
    \centering
    \includegraphics[width=0.95\linewidth]{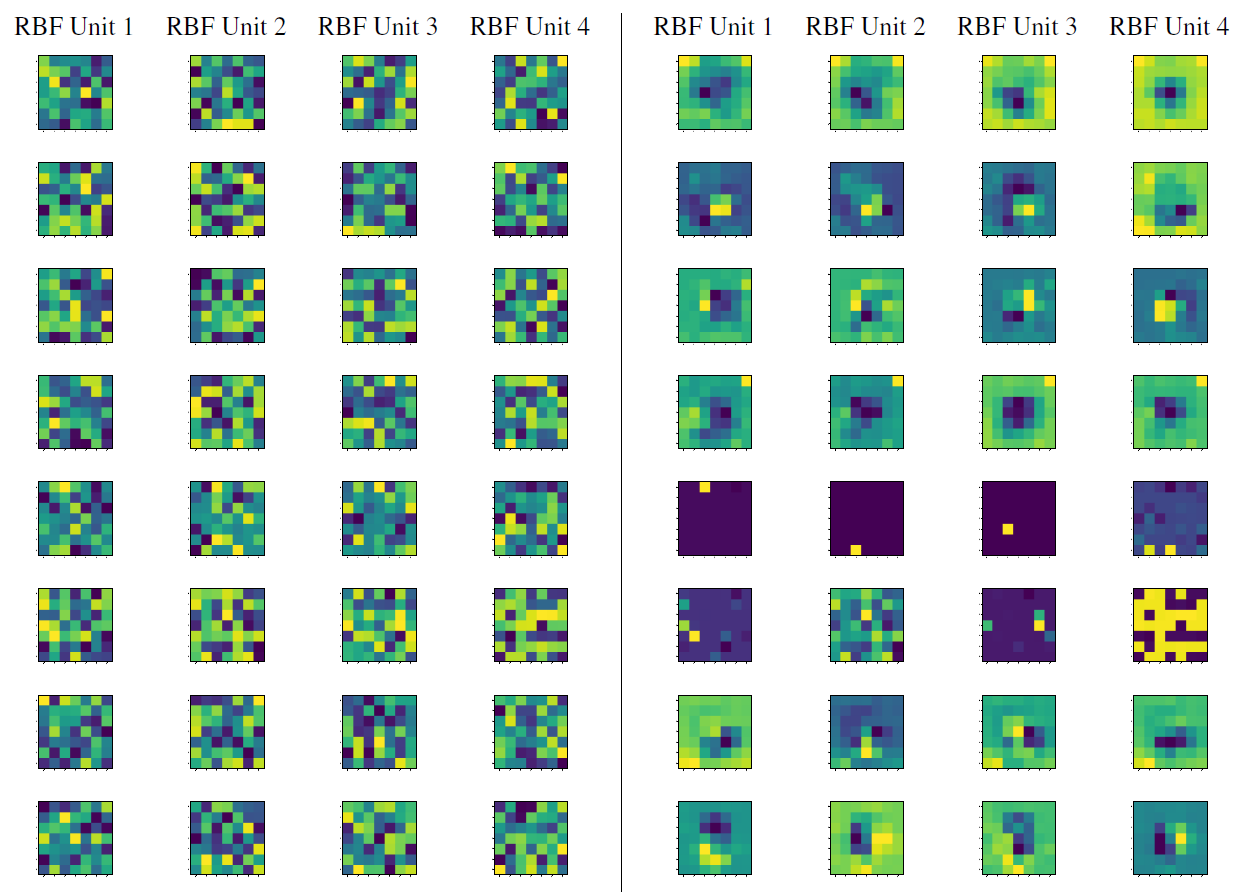}
    \caption{Visualization of the centers of the RBF units for a model with 8 branches and 4 RBF units. The left plot shows the initialized centers and the right one shows the learned centers after convergence.  }
    \label{fig:4units}
\end{figure}  

Figure~\ref{fig:4units} shows the centers but for a different configuration: 8 branches and 4 RBF units each, with a reported performance of 0.953. Again, centers seem to be visually different to each other, although the differences across RBF units of the same branch (rows, right plot) are less notorious. This could be reflecting the fact that branches are capturing local patterns with subtle differences across RBF units (except branch 5, fifth row in Figure~\ref{fig:4units} that seems to be learning the same pattern in the 3 RBF units).  In fact this type of centers result in better performance for the approached dataset. 
Finally, it is worth to emphasize that in both cases the centers seem to converge to an useful representation, starting from random numbers (left plots in Figures~\ref{fig:4ramas} and~\ref{fig:4units}).

\subsection{Comparison with reference models}
\label{sec:resultsbaselines}
Table~\ref{tab:results} shows a comparison of performance of the proposed model with other variants\footnote{Please note that in  preliminary experimentation other variants of CNNs were also evaluated (including the model used in~\cite{zadeh2018deep} and other configurations of CNNs that relied on different fine tuning processes). However, we are reporting only the most competitive baselines for comparison. } of CNN that approach the same task. We report the average and standard deviation obtained from 10 experiments with different random initialization. 
As reference models we considered: (1) the backbone model, VGG-Face, the pre-trained network was subject of a fine tinning process with the new classes, the last layer was removed and replaced by a softmax one with as many units as ER classes; the fully connected layers were re initialized and subject to the fine tuning process too. (2) Multi-branch CNN is a model in which the branches of RBF units are replaced by dense layers (see Figure~\ref{fig:Branch_Mod}). The idea is to determine whether adding parameters to the backbone is the cause of improvement. Overall, the goal of this experiment is to assess the benefits of the proposed model when compared to competitive models that do not incorporate local information.

\begin{table}[h]
\centering
\caption{Top-1 accuracy \textit{classification} for the considered datasets. We compare the performance of the proposed model (MB-RBFN) with reference models (VGG-Face and MB-CNN).}\footnotesize{
\begin{tabular}{|c|l|l|l|}
\hline
\textbf{Datasets}  & \textbf{VGG-Face} & \textbf{MB-CNN} & \textbf{MB-RBFN} \\ \hline
\textbf{CK+}      & 0.8291 $\pm$ 0.003   & 0.8381 $\pm$ 0.051          & \textbf{0.9964} $\pm$ 0.0037\\ 
\textbf{JAFFE}    & 0.6352  $\pm$ 0.012   & 0.5971 $\pm$ 0.032          & \textbf{0.9796} $\pm$ 0.0314\\ 
\textbf{CK+-JA} & 0.8341 $\pm$ 0.0018  & 0.8594 $\pm$ 0.021          & \textbf{0.9872} $\pm$ 0.0024\\ 
\textbf{FER13} & 0.4731 $\pm$ 0.035 & 0.6751 $\pm$ 0.0082 & \textbf{0.6815} $\pm$ 0.0097 \\
\textbf{RAF} & 0.4289  $\pm$ 0.058  & 0.7237 $\pm$ 0.041    & \textbf{0.810} $\pm$ 0.0014  \\ 
\textbf{RAF-C} & 0.2330 $\pm$ 0.0012  & 0.4739 $\pm$ 0.0034   & \textbf{0.5768} $\pm$ 0.0074                \\ \hline
\end{tabular}}
\label{tab:results}
\end{table}

From Table~\ref{tab:results} \textbf{\emph{it is clear that the proposed model outperforms both of the reference models.}} The differences in performance are significant for most datasets and there are also dramatic improvements in some cases. Compare for instance the performance of VGG-Face and the proposed model for the RAF-DB and RAF-DB Compound datasets. The differences in performance are impressive.
This could be due to the mismatch between the datasets (both in terms type of images and classes) used for training VGG-Face and the ones considered for evaluation, even when we fine tuned the FC layers of the model, the mismatch seems to be too large as to be learned by the FC layers.  Actually the MB-CNN baseline outperforms VGG-Face in all but the JAFFE dataset. Showing evidence that the added layers to the \emph{standard} VGG-Face architecture improved the recognition performance. 

\begin{figure}[h!tb]
    \centering
    \includegraphics[width=0.8\linewidth]{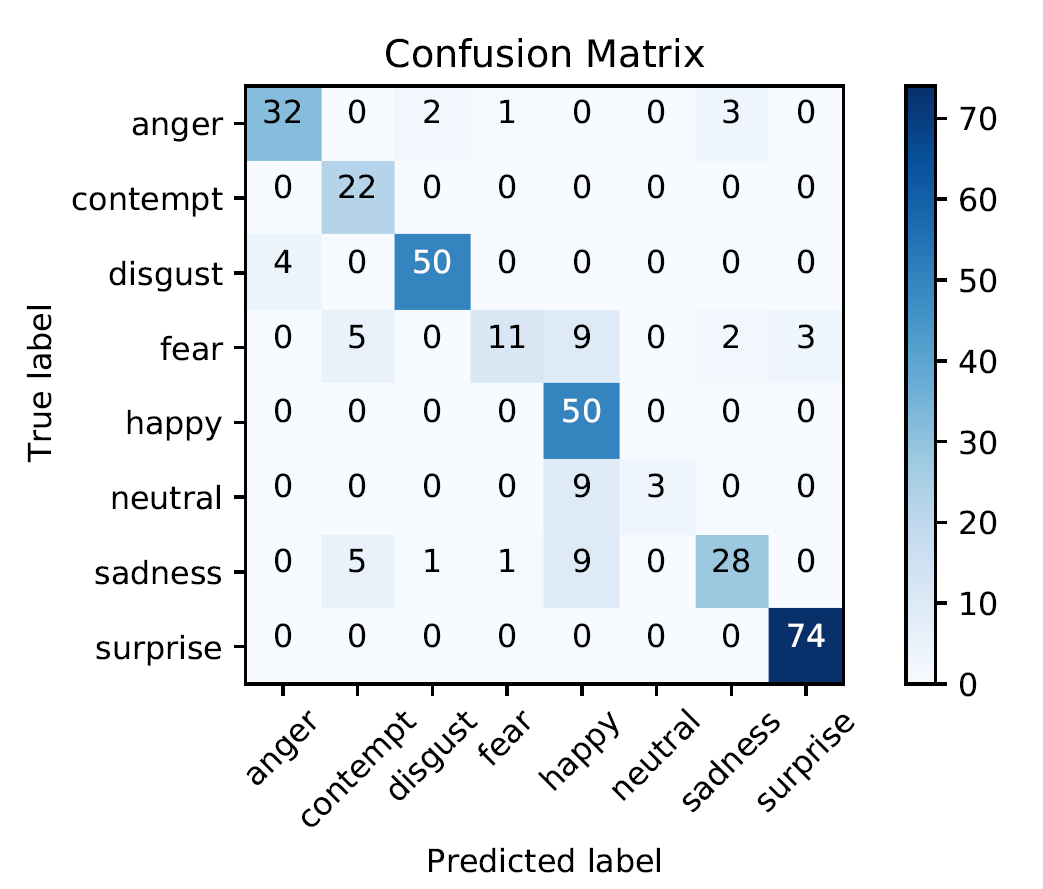}
        \includegraphics[width=0.85\linewidth]{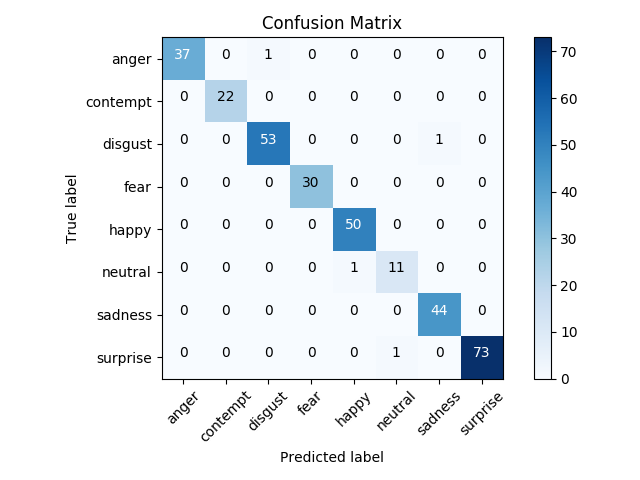}
    \caption{Confusion matrices obtained by VGG-Face (top) and the proposed model (down) for the CK+-JAFFE dataset.   }
    \label{fig:confmats}
\end{figure}

We further analyze the differences in performance between VGG-Face and MN-RBFN. 
Figure~\ref{fig:confmats} shows the confusion matrices for VGG-Face and the proposed model on the challenging CK+- JAFFE dataset. It can be seen that VGG-Face makes considerably more mistakes for the \emph{anger, fear} and \emph{sadness} categories. \textcolor{black}{Our model miss classified 4 images from the JAFFE dataset and 6 from CK+, whereas the VGG-Face model made 28 and 29 mistakes for JAFFE and CK+ images, respectively. This represents 58\% of images from JAFFE in the test set and only 10\% if CK+ images. This clearly illustrates the benefits of incorporating local information into the CNN model: \textbf{\emph{underrepresented samples are better classified}}} (similar behavior was observed for the other baseline model). We refer the reader to Appendix~\ref{ap:confmat} for a comparison of confusion matrices for the three models in the RAF-DB Compound dataset. 

\begin{figure}[h!tb]
    \centering
    \begin{tabular}{c}
    \subfloat[Surprise]{\includegraphics[width=0.9\linewidth]{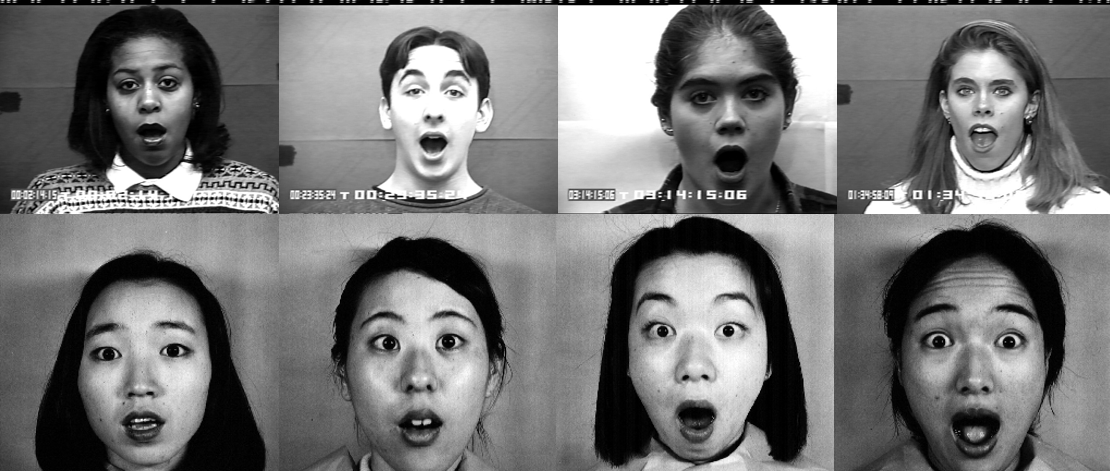}}\\
    \subfloat[Fear]{\includegraphics[width=0.9\linewidth]{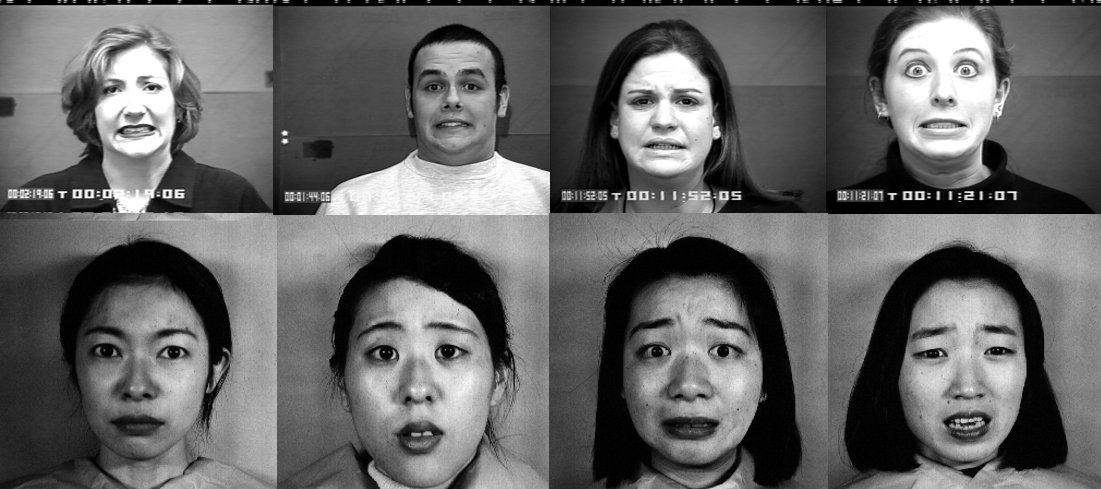}}%&
    %\subfloat[Sadness]{\includegraphics[width=0.9\linewidth]{sadness.png}}
    \end{tabular}
    \caption{Images from the test set of CK+-JAFFE for the \emph{surprise} and \emph{fear} categories. In each sub-figure the top row displays the CK+ images and the bottom one shows the JAFFE images.  }
    \label{fig:erroresckj}
\end{figure}

In order to further analyze these errors, Figure~\ref{fig:erroresckj} shows sample images from the \emph{surprise} and \emph{fear} categories. The former being one of the best classified by both models\footnote{We included images for \emph{surprise} instead of \emph{happy} because for the latter category an single test image from JAFFE was included, hence it is not an informative class.} and the latter the most difficult class for the VGG-Face model.  It can be seen from this figure that samples for the \emph{surprise} category share a notable pattern regardless of their origin: the mouth is open in all cases, this makes the \emph{generic} VGG-Face model to correctly classify most of the test instances in the mixed dataset. However, for the \emph{fear} category images coming from CK+ and JAFFE look visually  different to each other, yet sharing similarities within each dataset. This makes this class particularly challenging to VGG-Face, while the proposed model is able to correctly classify every instance from this class. This could be due to the local information incorporated into the model, and we think this is the main distinctive feature of our proposal. 

\textbf{\emph{Regarding the MB-CNN baseline (column 3 in Table~\ref{tab:results}), it is also outperformed by the proposed model in every considered dataset}}, where the lowest improvement obtained was for the FER2013 dataset. We hypothesize this could be due to the large number of images available for training in this dataset (more than 28,000), that allow the MB-CNN model to find a competitive configuration of parameters with the extended dense layer added to the VGG-Face model. Still, the proposed model obtained the highest performance overall (see Appendix~\ref{ap:confmat}). This difference in performance shows that is the \textbf{\emph{local information, as captured by the proposed model, was the decisive factor for obtaining better performance across the considered datasets}}. 
\subsection{Comparison with the state of the art}
\label{sec:sota}
In this section we compare the performance obtained by the proposed model with state-of-the-art references that have used the same datasets. Table~\ref{tab:sota} shows the results of the comparison. For each of the considered datasets we report the performance of recent references including the best result reported so far in each dataset to the best of our knowledge. One should note that for our model we report the average over 10 runs as reported in Table~\ref{tab:results}, while for the reference models we take the single best result in the corresponding references. We include the results obtained by the baseline models for completion.  
\begin{table*}[h]
\centering
\caption{Top-1 accuracy \textit{classification} for the considered datasets. We compare the performance of the proposed model with state-of-the-art references. The number between parenthesis is the position in the rank of the results when ranking performance from highest to lowest. }
\begin{tabular}{|c|l|c|l|c|l|c|l|c|l|}
\hline
\multicolumn{2}{|c|}{\textbf{CK+}}&\multicolumn{2}{|c|}{\textbf{JAFFE}}&\multicolumn{2}{|c|}{\textbf{FER2013}}&\multicolumn{2}{|c|}{\textbf{RAF-DB}}&\multicolumn{2}{|c|}{\textbf{RAF-DB C}}\\\hline
Ref.&Acc.&Ref.&Acc.&Ref.&Acc.&Ref.&Acc.&Ref.&Acc.\\\hline
\cite{chen2019facial}&0.9806&\cite{li2020attention}&\textbf{0.9852}&\cite{liang2020fine} & \textbf{0.7830}&  \cite{wang2020suppressing} & \textbf{0.8814}&\cite{li2019separate} & \textbf{0.5884}\\
%\cite{s20174727}&\textbf{0.6697}\\
\cite{minaee2019deep}&0.9800&\cite{shima2018image}&0.9531&\cite{li2020attention} & 0.7582&\cite{farzanehfacial}   &  0.8778& %\cite{li2017reliable}  & 0.5354 (3)\\
%\cite{li2019separate} & 0.5884 \\
\cite{li2017reliable}  & 0.5795\\
\cite{ravi2020face} & 0.9732&\cite{minaee2019deep}&0.9280&\cite{minaee2019deep} & 0.7002& \cite{zeng2018facial} & 0.8690& \cite{li2017reliable}  & 0.5354 \\ %****
\cite{wang2020region} & 0.9730&\cite{zhang2020facial} & 0.9238& \cite{mollahosseini2016going} & 0.6640&\cite{zhao2016peak}    & 0.8677&\cite{liang2020fine} &0.5020\\
\cite{meng2017identity}&0.9537& \cite{videla2020facial} & 0.7810&\cite{shao2021fcnn} & 0.6617&\cite{liang2020fine} & 0.758 &\cite{DBLP:journals/corr/abs-1910-11111}&0.4830\\%\hline
VGG-Face&0.8295&&0.6352&&0.4731&&0.4289&-&0.2330\\
MB-CNN&0.8381&&0.5971&&0.6751&&0.7237&-&0.4739\\\hline
MB-RBF&\textbf{0.9964}&&0.9796&&0.6815&&0.81&-&0.5758\\\hline
\end{tabular}
\label{tab:sota}
\end{table*}

From Table~\ref{tab:sota} it can be seen that it is only in two datasets, FER2013 and RAF-DB, out of the five considered for this evaluation that the proposed model does not achieve performance competitive with the state-of-the-art. Interestingly,  these are precisely the two datasets with the largest number of samples with 28,807 and 12,271 respectively. This result seems to indicate that the proposed model is particularly helpful for low-mid sized datasets. Likewise, since the references under comparison are based on extremely complex models and sophisticated procedures, it is not strange that they perform better when enough data is available. 

\begin{figure}[h!tb]
    \centering
    \includegraphics[width=0.9\linewidth]{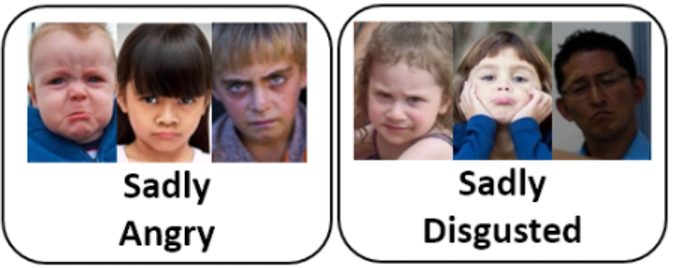}
    \caption{Sample images from two of the classes in the RAF-DB Compound dataset, image taken from \url{http://www.whdeng.cn/RAF/model1.html}.}
    \label{fig:samplerafc}
\end{figure}

On the other hand, the proposed model achieves very competitive performance in CK+, JAFFE and RAF-DB Compound datasets. In CK+  our work establishes a new reference result and in the JAFFE and RAF-DB Compound datasets the model achieves comparable performance. It is remarkable the performance obtained by the proposed model in the RAF-DB Compound dataset, as this  features a problem of (very) fine grained classification with overlap among classes (compare the classes \emph{Sadly-Disgusted} and \emph{Sadly-Angry}, see Figure~\ref{fig:samplerafc}) and highly imbalanced (4 classes comprise $72\%$ of the samples, and the 7 remaining classes with 6\% of less out of the total number of samples). This result provides further evidence that the model is particularly helpful for this type of problems. 

It is important to emphasize that among the  references considered in the comparison we are including all types of recent models and mechanisms and it is to some extend unfair to compare a model like ours, which uses a simple backbone (VGG-Face). For instance, \textcolor{black}{for JAFFE the only method that obtains better performance than our model is based on a complex CNN equipped with attention mechanisms and taking advantage of both learned and handcrafted features. } \textcolor{black}{Whereas for RAF-DB Compound, the method with the highest accuracy is a ResNet18 model (that has proven to be superior to VGG-Face) with a so called \emph{separate loss} that enhances the initial architecture to consider intra and inter class information for the ER process.}  
%an ensemble of ResNET50 (that has proven to be very superior to VGG-Face) with extended channel inputs and DenseSift descriptors further processed with a LSTM network. The model is pretrained in the whole RAF-DB dataset.
Clearly, \textbf{\emph{our model is advantageous in terms of simplicity, besides, it is possible that if we rely on a more complex backbone models the performance of the multi-branch RBF model could be even superior}}. Finally, please note that our model is not doing any \emph{ad hoc} feature learning process: we are relying on the pre-trained VGG-Face model, while most other references comprise expensive learning-from-scratch or fine-tuning processes that often use additional external data. 

\subsection{Discussion}

We have presented an experimental evaluation of the proposed \emph{Multi-Branch Deep RBF} Network model. We reported experiments in six datasets widely used for ER, two of them were  variants that presented particular challenges with which state-of-the-art methods struggle. Our experimental evaluation showed that the proposed model outperforms considerably to reference models that included a similar model formed by dense layers only and the backbone. The comparison with the reference models together with a visual inspection of the learned centers comprises evidence that local information as captured by the proposed model is useful for approaching the ER task. 

On the other hand, the proposed model compared favorably with recent methodologies that are based on much more complex techniques and procedures.  This is an outstanding result given that the proposed model relies on a very generic, yet effective, backbone model: VGG-Face. Interestingly, it was shown that the proposed model offers is more advantageous in datasets with more challenging conditions, namely: small-medium sample size, with high class-imbalance, class overlap and with images coming from two different distributions.   
The obtained results are thus encouraging and motivate further research on the incorporation of local information into deep learning.

\section{Conclusions}
\label{sec:conclusions}
We introduced the \emph{Multi-Branch Deep RBF} Network, a model that improves CNNs by a mechanism that allow it to incorporate local information in the recognition process.  The proposed model relies on VGG-Face as backbone for feature extraction, where the last convolutional layer of this model is connected to multiple branches of RBF units. The outputs of these are concatenated and connected to a softmax layer. The proposed model is initialized with VGG-Face and the RBF layers are fine tuned. Experimental results are reported in six ER datasets.  

The following summarize the main findings of this work:
\begin{itemize}
    \item The inclusion of local information, via the multi-branch RBF units, improves significantly the performance of a CNN model. In fact, the proposed model outperforms a similar model extended with a dense layer, showing that the RBF units are  responsible of the improvement in performance. 
    \item The proposed model is competitive with state-of-the-art methods based on more complex architectures and mechanisms, even when we rely on a standard backbone (VGG-Face). The model achieved competitive results in 3 out of 5 datasets with a much simpler implementation. 
    \item The proposed model proved to be more advantageous for datasets with challenging conditions that include small sample size, high overlap among classes, datasets with mixed distributions in the test set and with high imbalance ratios. 
    \item The centers of  RBF units from different branches capture local information and this information resulted very helpful for classifying samples coming from different distributions (e.g., with the CK+-JAFFE dataset). 
\end{itemize}

The findings and results presented in this paper are encouraging and motivate further research. In particular, in future work we would like to explore alternative ways of incorporating local information in deep learning based models. Likewise, we are analyzing the ways in which information from the RBF centers, in the proposed model, can be used for explainability and interpretability. Finally, another exciting research direction is to extend the proposed method so that it can be used to undercover biases in FER. 

\bibliographystyle{plain}
\bibliography{References}   % name your BibTeX data base
\appendix
\section{Confusion matrices for RAF-DB Compound}
\label{ap:confmat}
\begin{figure}[]
    \centering
    \includegraphics[width=0.47\linewidth]{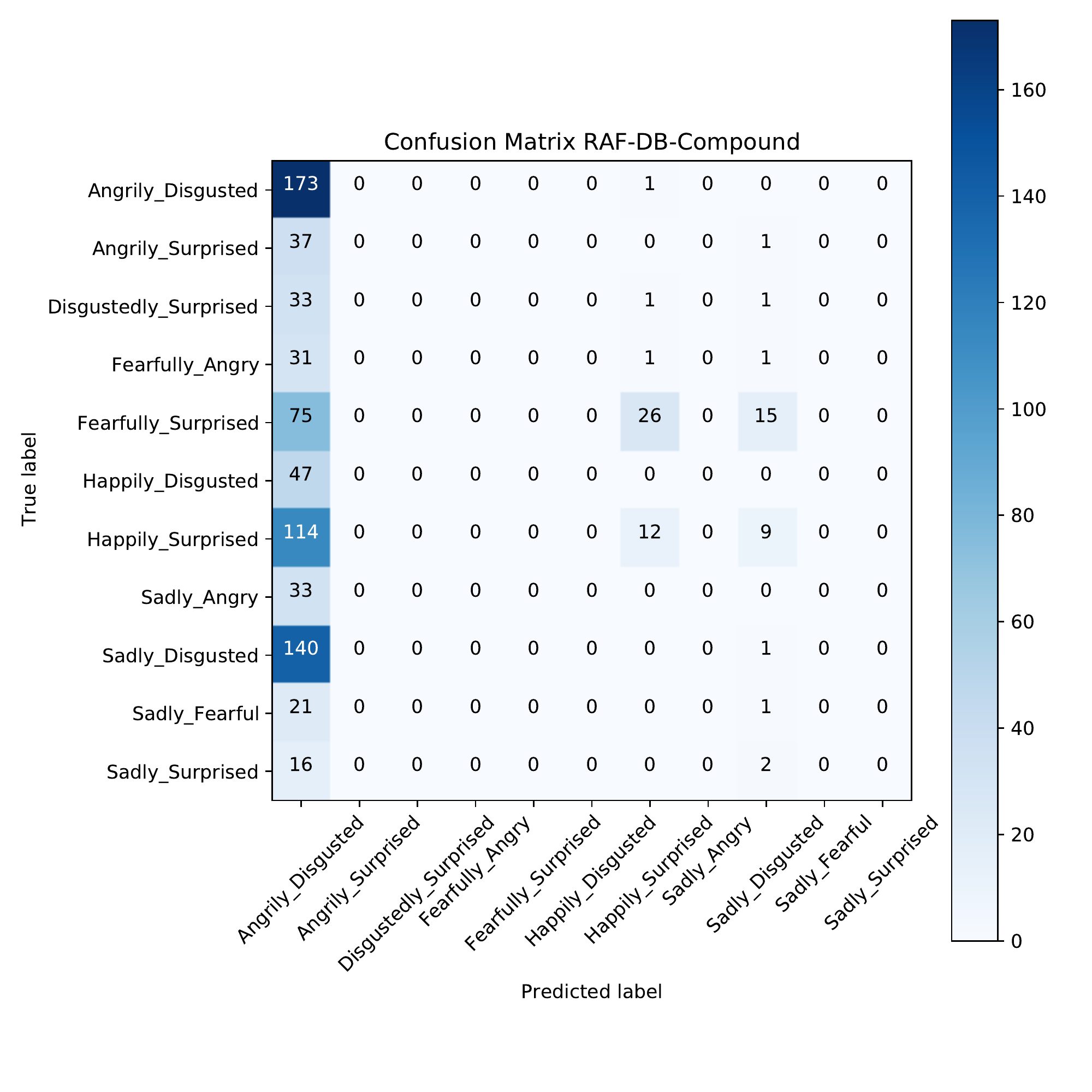}
        \includegraphics[width=0.47\linewidth]{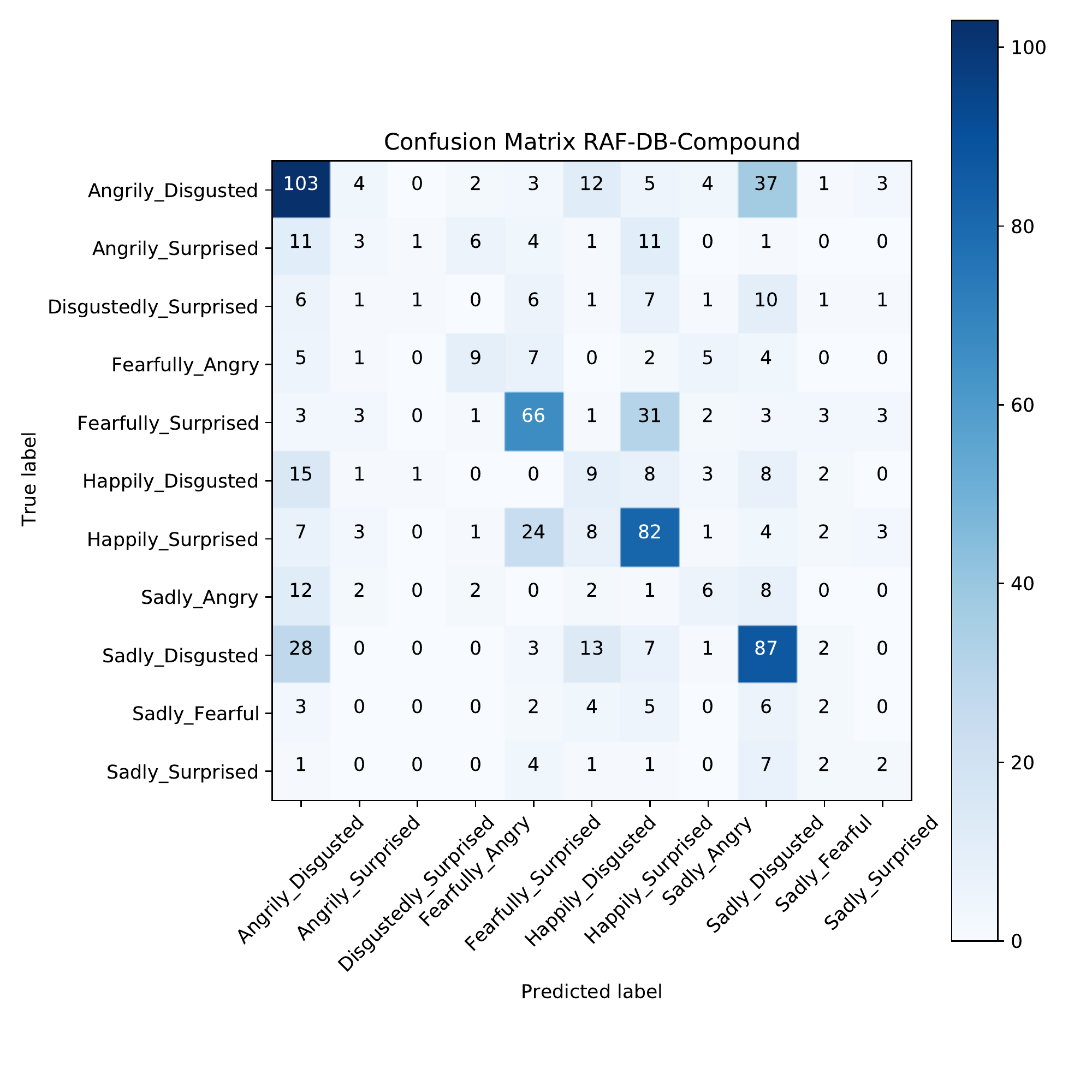}
        \includegraphics[width=0.47\linewidth]{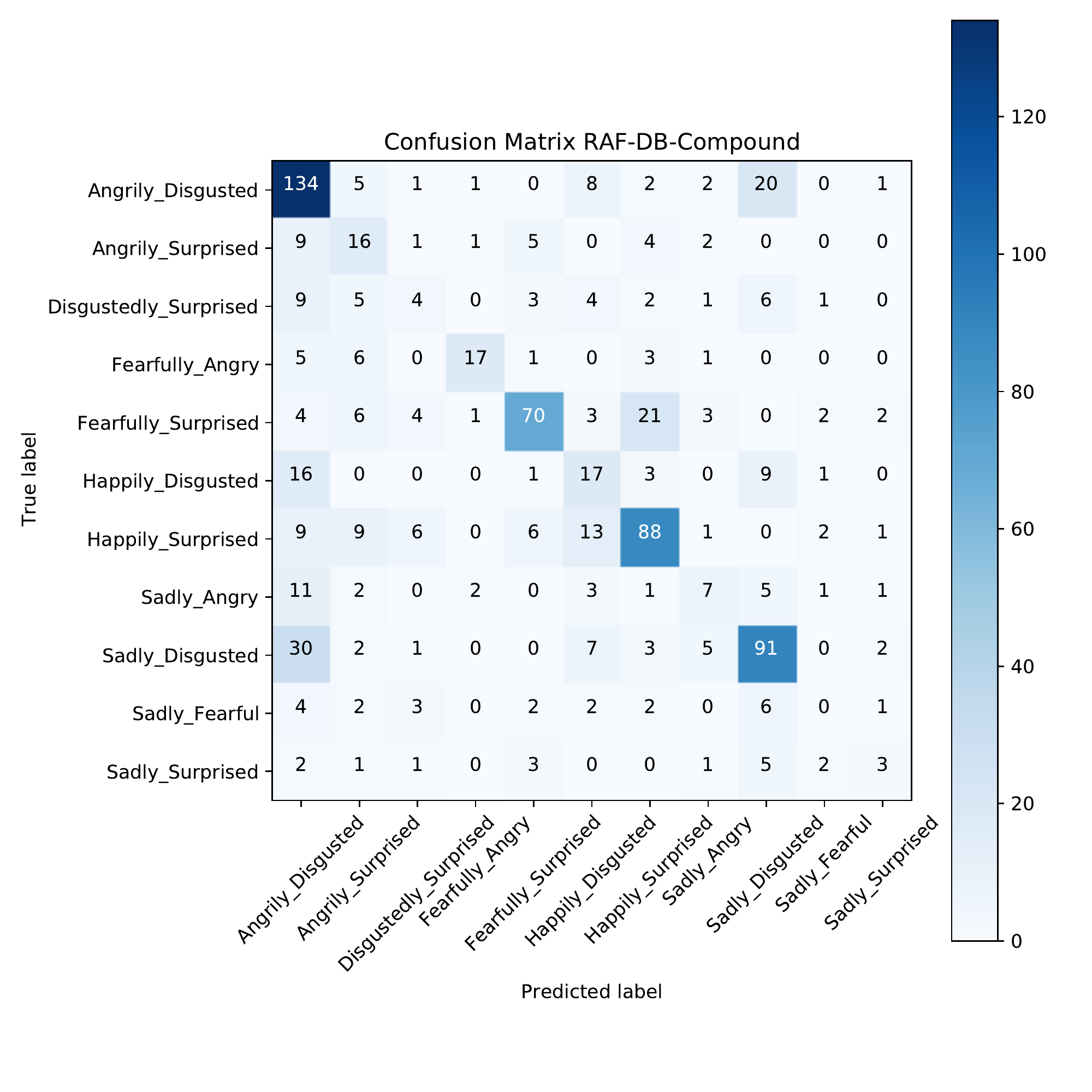}
    \caption{Confusion matrices obtained by VGG-Face (top) MB-CNN (middle) and the proposed model (down) for the RAF-DB-Compound dataset.   }
    \label{fig:confmats2}
\end{figure}

In this section we analyze the confusion matrices for the proposed and reference models in the RAF-DB Compound dataset. Figure~\ref{fig:confmats2} shows these confusion matrices. From this Figure it can be seen that the proposed model is advantageous over both reference models. Consider, for instance, the \emph{Angrily-Disgusted} class: despite this is the majority class, it is a very difficult to predict for the three models; clearly, MB-RBF obtains the best performance, by reducing considerably the number of miss classifications made by VGG-Face and MB-CNN.  Similar behavior can be observed for non majority classes. This analysis illustrates the benefits offered by incorporating local information into the proposed MB-RBF model. 
%Interestingly, the proposed model improves considerably the recognition performance for minority class when compared to the performance of the baselines. Compare the \emph{Sadly-Disgusted} and \emph{Sadly-Angry}
\end{document}